\documentclass[letterpaper, 10pt, conference]{ieeeconf}      

\IEEEoverridecommandlockouts                              

\usepackage{graphicx}
\usepackage{algorithm}
\usepackage[noend]{algpseudocode}
\usepackage{amsmath}
\usepackage{nicematrix} 
\usepackage{multirow}  
\usepackage{comment}
\usepackage{float} 

\usepackage{tikz, pgfplots, fix-cm}
\usetikzlibrary{plotmarks}
\usepackage{graphicx,booktabs,array}

\usepackage{amssymb}
\usepackage{pifont}
\def\secref#1{Sec.~\ref{#1}}
\def\figref#1{Fig.~\ref{#1}}
\def\tabref#1{Tab.~\ref{#1}}
\def\eqref#1{Eq.~(\ref{#1})}

\newcommand{\RNum}[1]{(\uppercase\expandafter{\romannumeral #1\relax})}

\title{\LARGE \bf BonnBot-I: A Precise Weed Management and Crop Monitoring Platform}

\author{Alireza Ahmadi, Michael Halstead, and Chris McCool
	\thanks{All authors are with the University of Bonn, Bonn 53115 Germany. 
			{\tt\small \{alireza.ahmadi, michael.halstead, cmccool\}@uni-bonn.de}}%
}

\newcommand\etal{\emph{et al.}}
\newcommand\bbot{BonnBot-I}

\newcommand\TODO[1]{\textbf{\textcolor{red}{#1}}}

\begin{document}

\maketitle
\thispagestyle{empty}
\pagestyle{empty}

\begin{abstract}

Cultivation and weeding are two of the primary tasks performed by farmers today.
A recent challenge for weeding is the desire to reduce herbicide and pesticide treatments while maintaining crop quality and quantity.
In this paper we introduce \bbot\ a precise weed management platform which can also performs field monitoring.
Driven by crop monitoring approaches which can accurately locate and classify plants (weed and crop) we further improve their performance by fusing the platform available GNSS and wheel odometry.
This improves tracking accuracy of our crop monitoring approach from a normalized average error of $8.3\%$ to $3.5\%$, evaluated on a new publicly available corn dataset.
We also present a novel arrangement of weeding tools mounted on linear actuators evaluated in simulated environments.
We replicate weed distributions from a real field, using the results from our monitoring approach, and show the validity of our work-space division techniques which require significantly less movement (a $50\%$ reduction) to achieve similar results.
Overall, \bbot\ is a significant step forward in precise weed management with a novel method of selectively spraying and controlling weeds in an arable field.

\textit{Keywords} — Robotics and Automation in Agriculture and Forestry; Agricultural Automation; Field Robotics.

\end{abstract}

\section{Introduction}
\label{sec:indroduction}
The preference for consuming more natural and organic foods has rapidly increased in recent years~\cite{blasco2002ae}.
This has forced the agricultural industry to use fewer agri-chemicals when dealing with weeds while maintaining the quality and quantity of the crop.
Weed intervention is an important aspect of arable farming due to its competition with crops for nutrients in the soil~\cite{slaughter2008autonomous} which potentially reduces yield.
Currently, to alleviate this impact, the majority of farmers use uniform treatments for weed control for instance by treating the entire field with herbicide irrespective of weed presence. 
This approach has led to an increasing number of herbicide-resistant weed species~\cite{heap22_website} as well as negatively impacting the environment by increasing soil erosion and water contamination~\cite{mia2020sustainable}.

Robotic weed intervention has the potential to revolutionize weeding paradigms through plant-level weed management.
For instance, by only treating a plant if it is present and using the most appropriate action for the particular plant species~\cite{Bawden17_1}.
To achieve plant and species specific treatments, robots are driven by advanced perception systems that can also provide rich crop monitoring information~\cite{halstead2021crop}.
Yet, robotic design has primarily explored how to design the platform to perform weeding operation.

Several robotic weed control platforms have been introduced offering active and passive interventions in field.
A variety of weeding implements have been investigated including physical~\cite{chang2021mechanical}, chemical~\cite{wu2020robotic}, electrocuting~\cite{ascard200710}, laser-based~\cite{xiong2017development}.
Given the variety of tools, it is clear that there is no one best solution and robotic solutions should be able to cater to a variety of tools.
The multi-modal approach of Bawden et al.~\cite{Bawden17_1} provided a clear step in this direction, however, a downside of their approach was the need to densely replicate each tool as they were mounted statically.
We propose to overcome this limitation by considering replicated movable tools.
 
\begin{figure}[t]
	\centering
	\includegraphics[width=1.0\linewidth]{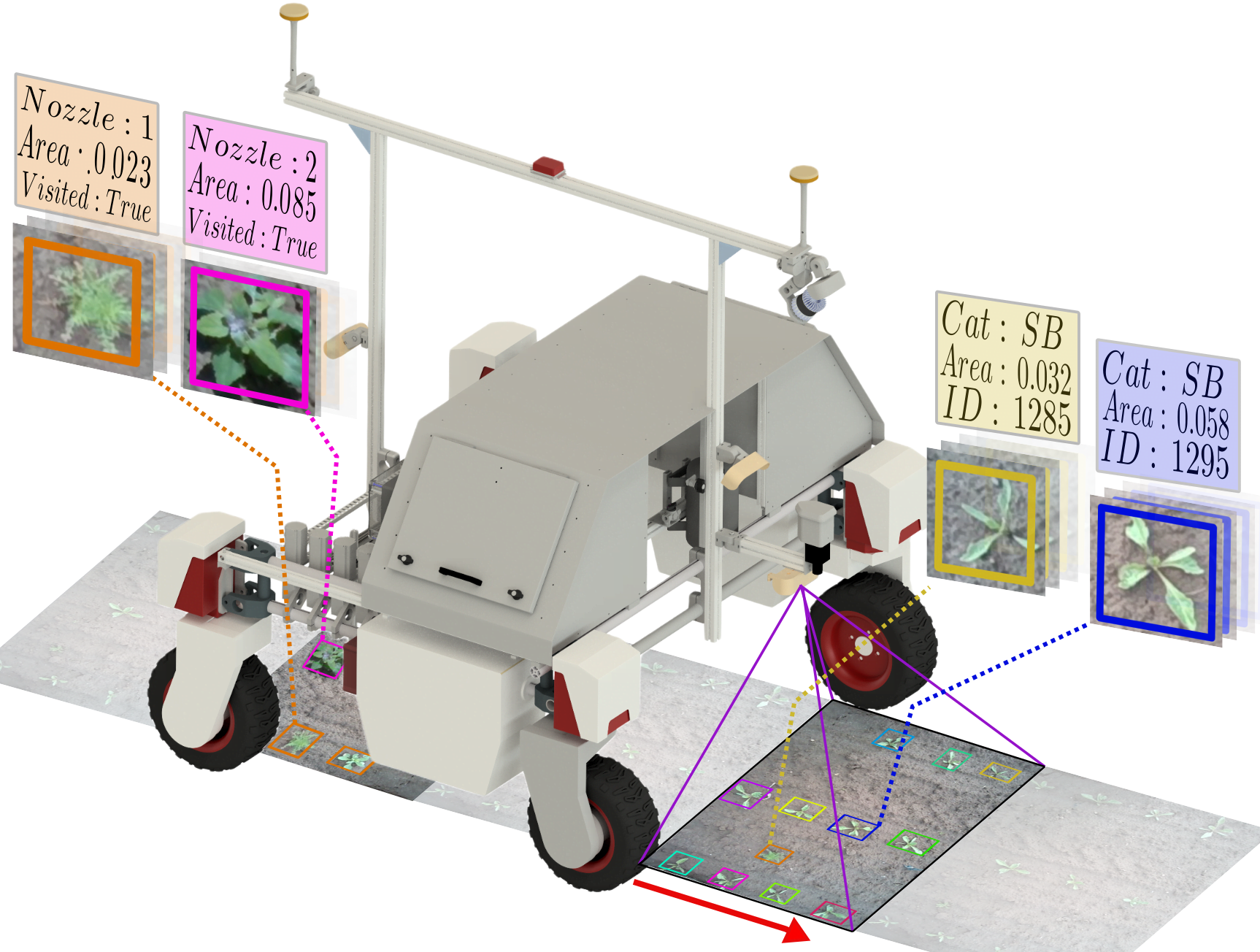}
	\caption{\bbot\ Platform, a robotic platform capable of conducting field monitoring and precision weed management in arable field.}
	\label{fig:motivation}
	\vspace{-7mm}
\end{figure}

In this paper we introduce \bbot, a robot that performs both field monitoring and precision weed management.
It enhances the capabilities of a recently published field monitoring technique~\cite{halstead2021crop} by using multiple localization sensors (GPS and odometry).
The plant counting performance is improved, reducing the normalized absolute error by more than half from $8.3\%$ down to $3.5\%$.
In developing \bbot\ we also propose a novel arrangement of weeding tools to enable precise weed management by replicating linear systems.
An advantage of this approach is that we can deploy both fewer and smaller (more precise) tools, rather than densely replicated tools~\cite{Bawden17_1}, while still allowing for good coverage.
We demonstrate that it is feasible to have a system with just 4 replicated tools of size $0.05m$ on linear systems to cover a width of $1.3m$ which would normally require at least $26$ non-overlapping tools.
This leads to the following novel contributions:
\begin{enumerate}
    \item We introduce BonnBot-I a fully autonomous precision weeding platform fully compatible with ROS.
    \item Propose a new concept for weeding tools that enables flexible high-precision weed management.
    \item Improve crop monitoring performance by exploiting the extra sensors available with \bbot. 
    \item Release a new dataset consisting of corn as the crop, CN20. 
    This is a challenging dataset for crop monitoring approaches as it is a grass crop. 
    \item Introduce a framework for testing different weeding intervention strategies using a simulation environment.
\end{enumerate}

\section{Related Work}
\label{sec:relatedworks}

Robotic platform based weed management techniques have rapidly evolved in the last decade with the aim to treat each weed as precisely as possible. 
To achieve this, plant-level intervention needs to operate in different fields with varying crops, weed species and weed distributions.
To enable this, ``smart farming techniques'' aim to incorporate automated navigation~\cite{ahmadi2021towards}, crop monitoring~\cite{halstead2021crop}, and weeding~\cite{pretto2020building}. 
%
%
%
One of the key elements to achieve precise weeding is plant-level treatment, where the treatment of each plant is dictated by its species, size and its impact upon not only the crop but also the environment~\cite{blaix2018quantification}.
%
%
%
Clearly, these approaches rely on the underlying perception or agricultural monitoring approaches which have gained significant research attention in recent years, including in glasshouses~\cite{smitt2021pathobot, smitt2022explicitly}, orchards~\cite{tian2019apple}, and fields (for weed intervention)~\cite{peruzzi2017machines}.

To achieve precise species level intervention recent methods have relied on deep learning due to its accuracy and generalizability.
Tian et~al.~\cite{tian2019apple} compared both Faster-RCNN~\cite{ren2015faster} and Yolo~\cite{redmon2018yolov3} and found Yolo to be superior in both performance and speed, while,~\cite{wan2020faster} was able increase both the performance and speed of Faster-RCNN.
Expanding on~\cite{halstead2020},~\cite{halstead2021crop} showed the viability of crop agnostic monitoring by exploiting an augmented Mask-RCNN framework. 
To improve the accuracy of monitoring techniques in arable farmland~\cite{ahmadi2021virtual}, was able to utilize the temporal information captured by the platform.
To selectively weed~\cite{pretto2020building} built, based on the work in~\cite{lottes2018fully}, a fully convolutional network that was able to accurately classify weeds in the field.
In order to perform intervention platforms are required, in recent years the number of available robotic platforms for weed management has increased.
An early, low cost, platform was developed by~\cite{Ruiz14_intracrop_weeding} which contained a mechanical weeding tool for intra-row intervention that required a human in the loop.
Bawden~\etal~\cite{Bawden17_1} proposed an automated platform that utilised a row of weeding hoes and spray nozzles to improve on broadcast applications and perform multi-modal weeding (physical and chemical).
Also combining mechanical tools (two ranks of stampers) and sprayers~\cite{pretto2020building} is able to selectively weed based on the overall size of the plant.
To design a spraying platform~\cite{zhou2021design} developed a technique that was able to specifically target regions or weeds. 
In an indoor experiment they were able to reduce the amount of chemicals used, compared to a uniform sprayer, by $46.8\%$.
Chang~\etal~\cite{Chang21_mechanical_tools} built a platform to evaluate two different mechanical weeding tools which was tested on a purpose built field with a single crop-row of 20m.
Their approach used Yolo to both locate weeds of interest in the field but also control the movement of the platform.

A consistent trend with the above approaches is that they aim to improve on broadcast or uniform weeding by introducing equally spaced tools.
In doing so they ensure coverage but have the significant downside of having to replicate the tool across the entire width of the robot.
Yet, the optimal tool is dependent upon the field and plants as shown by the variety of tools: physical~\cite{chang2021mechanical}, chemical~\cite{wu2020robotic}, electrocuting~\cite{ascard200710} or laser-based~\cite{xiong2017development} implements.
As such, we conclude that to provide truly precise intervention a weeding platform needs the ability to mount multiple tools.
This issue exacerbates the problem of having replicating equally spaced tools as two or more sets of tools have to be replicated.
An alternative is to have movable tools on a robot, but this is only possible with an appropriate planning algorithm.

A frequently overlooked aspect for weeding is planning field based intervention.
If a robot carries multiple movable tools then planning their deployment is essential, yet limited work has explored this aspect.
Lee~\etal~\cite{lee2014fast} is one of the few works in this area.
They presented a multi-query approach for efficiently planning paths using a single UR5 robot manipulator to enable precision weeding.
This was achieved by maintaining a database of pre-computed paths that was constructed offline and the optimal path was chosen and adapted online.
Xion~\etal~\cite{xiong2017development} proposed a laser-based intervention scheme that segmented the visible weeds into equally spaced regions along the x-axis, this then allowed them to engage targets sequentially.


In this work, we aim to alleviate some of the limitations of the previous approaches by introducing a set of movable and replicated tools on a single weeding platform.
Furthermore, we present a robot that is able to navigate down European standard crop planting patterns which automatically monitors the crop with species level classification and localization.
This provides both automated field monitoring and weed management capabilities.

\begin{figure}[!t]
    \centering
    \vspace{2mm}
    \includegraphics[width=1.0\columnwidth]{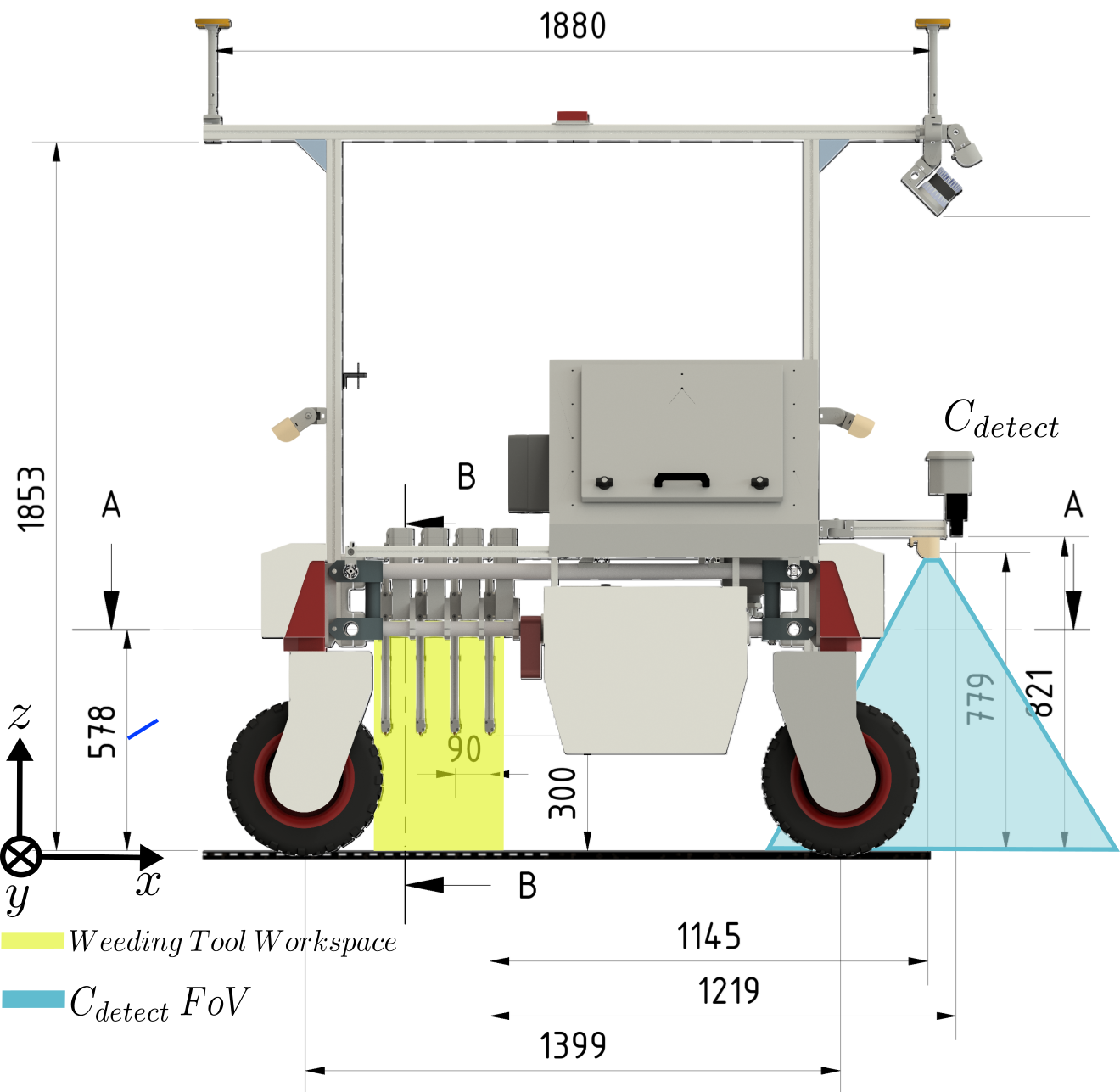}
    \vspace{-4mm}
    \caption{Overview of \bbot\ with dimensions of key aspects such as height, clearance, camera and tool placement (all numbers in centimeters).}
    \vspace{-7mm}
\label{fig:fields}
\end{figure}

\section{BonnBot-I}
\label{sec:platformHardware}

\bbot\ is a platform capable of both field monitoring and weed management developed at the University of Bonn.
The base platform is a Thorvald system~\cite{grimstad2017thorvald} which is a lightweight four-wheel-drive (4WD) and four-wheel-steering (4WS) system.
With considerable modifications, we adapted this platform to an arable farming and phenotyping robot suitable for operation in the real phenotyping fields.

These modifications were carried out to ensure they met European farming standards for the distance between rows.
Based on these specifications the width from wheel-centre-to-wheel-centre was set to $1.5m$.
To ensure we could monitor the different growth cycles of our primary crops (sugar beet, wheat, and corn) the vertical clearance is set to $0.57m$.
The length of BonnBot-I, $1.39m$, was selected to ensure there was adequate space to install replicating weeding tools.
These dimensions also act to increase the stability of the platform on the uneven terrain witnessed in arable farmland.
An overview of the \bbot\ platform is provided in~\figref{fig:fields}.

\subsection{Sensor Configuration}
\label{subsec:sensorConfiguration}

We equipped BonnBot-I with a range of sensors to perform both in-field weed management and crop monitoring.
There are two sets of sensors, the first set is for ``localization and navigation'' and the second set is for ``robotic vision''.

\subsubsection{Localization}
\label{subsubsec:navigation}

The localization of BonnBot-I is performed with a compact Inertial Navigation System (INS), Ellipse2-D SBG Systems~\cite{sbg} which includes an IMU and a dual-antenna receiver, multi-band GNSS receivers fixed at the front and back of the robot at height of $1.85m$ above the ground.
Using an on-board high-frequency extended-Kalman filter fusion of IMU and GPS data provides us with a horizontal and vertical position accuracy of $4cm$ and $3cm$ respectively. 
Furthermore, the heading of the platform can be determined with an accuracy of $0.1^{\circ}$ and $0.3^{\circ}$ in roll-pitch and yaw directions respectively.
%




\subsubsection{Robotic Vision}
\label{subsubsec:detectionInference}
BonnBot-I is equipped a nadir-view cameras (Intel Real-sense D455) on the front of the platform used for monitoring purposes.
The Real-sense D455 is a global shutter camera which provides RGB and registered depth images at $15 Hz$.
On \bbot\ it is fixed at a height of $0.78m$ providing a view-able area of $1.4m\times0.78m$ covering the gap between the two front wheels. 
This is the sensor used to perform field monitoring.


\subsection{Weeding Implements}
\label{subsec:weedingImplements}


Achieving flexible, and repeatable weeding implements that can deploy a variety of end-effectors is a key objective for BonnBot-I.
This enables the system to change tools given the current soil and weed populations.
The proposed design utilizes independently controlled high-resolution Igus linear actuators fixed at height of $0.72m$ above the ground and creating a working space of $1.3m\times0.36m$; the current design uses 4 such linear actuators.
Each linear axis has a length of $1.3m$ and is controlled by an Igus dryve D1 motor control system via Modbus connection with a maximum resolution of $0.01cm$ and is capable of performing translations with maximum velocity and acceleration of $5m/s$ and $10m/s^2$, respectively.
All linear axes are equally-spaced and currently carry spot-spray nozzles, however, the system design permits many kinds of end-effectors, such as mechanical hoeing, providing flexibility. 

To control the linear actuators and spray valves we use a ROS operated Raspberry-Pi 3B and to ensure minimal action delay the nozzles are accessed via high speed N-channel MOSFET-transistors.
Hence, ultimate operation time for each spray head in our system adds up to $10\sim12$ms including valves On-Off time. 
The spray system consists of a reservoir tank capable of carrying $5L$s of compressed liquid with a maximum $16bar$ pressure, as well as a compact $8bar$ portable compress which is fixed on the robot.
As the droplet size from the spray nozzles depends on the liquid pressure we use individually adjustable valves.
This allows us to control the spray footprints on the ground individually for each nozzle between $0.02m$ to $0.13m$; for this paper we assume a constant spray footprint of $0.05m$.

\subsection{In-Field Intervention Pipeline}
\label{subsec:interventionPipeline}
All sensors on \bbot\ are directly connected to a high performance fanless embedded computer (DS-1202) powered by 7th Generation Intel® Core™ processor running the robot operating system (ROS). 
This computer is equipped with an Nvidia® Quadro P2200 featuring a Pascal GPU with 1280 CUDA cores, 5 GB GDDR5X on-board memory which is used for parallel processing and inference of DNNs.
Additionally, a dedicated Intel® NUC PC runs the controller for the platforms motion providing wheel odometry data and the status of batteries and electronic infrastructure on the platform. 

The software architecture of \bbot\ contains four different nodes enable selective in-motion intervention.
\RNum{1} Field Monitoring: runs Mask-RCNN for instance-based semantic segmentation and intra-camera tracking which estimates necessary phenotypic information about the plants, more details are in~\secref{subsec:fieldMonitoring}.
\RNum{2} In-field localization: improves the localization accuracy of the robot using an EKF to fuse GPS and wheel odometry data~\cite{wei2011intelligent}.
\RNum{3} Intervention Controller: manages the targets within the work-space of the weeding tools and includes planning paths for the intervention heads (sprays) described in~\secref{subsec:selectivePreciseIntervenstion}.
\RNum{4} Weeding Implement: provides low-level control of the weeding tool (e.g. actuation) by taking commands from the intervention controller.
%
\subsection{Weeding Simulation Framework}
\label{subsec:simulation}
\begin{figure}[!t]
	\centering
	\vspace{2mm}
	\includegraphics[width=1.0\linewidth]{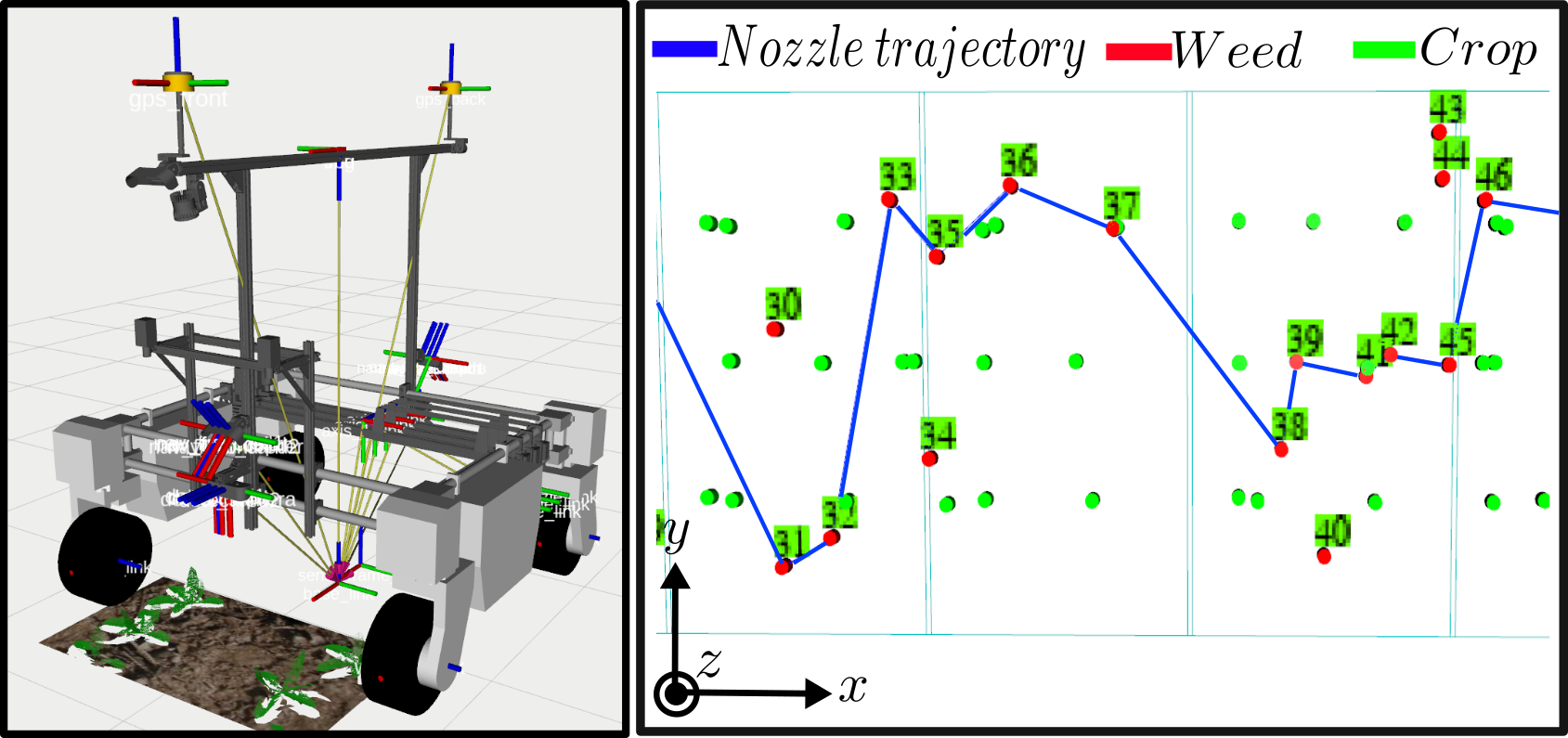}
	\caption{\bbot\ ROS simulation with active sensors and actuators (left), An example view of weeding simulation Framework (right) with green markers as plants and red markers representing weeds in each segment.
	A weeding path is also visualized in blue line for single axis weeding scenario.}
 	\vspace{-6mm}
	\label{fig:simulationModel}
\end{figure}

Conducting experiments in real fields is potentially time consuming and costly.
An accurate and reliable simulation environment that mimics real world situations is an invaluable tool which avoids this issue.
We use two different types of simulation environments for development and evaluation of our methods on \bbot\, a ROS based one-to-one scale simulation and a native python simulator especially developed for weeding application and intervention motoring purposes. 
A demonstration of the ROS-based simulation model is shown in~\figref{fig:simulationModel}-(left), where all sensors and actuators are active. 
Then a field generator allows us to create row-crop fields with different plant sizes and crop-row shapes, simulating a real field.
To evaluate the weeding algorithms, we developed a native Python based framework capable which simulates the robot kinematics and generates synthetic crop-rows with varying weed distributions.
This framework uses Open3D and Pyglet python libraries for rendering graphics, an simplified example view of the planning scenario is shown in~\figref{fig:simulationModel}-(right).
We used this environment to implement and evaluate different weeding strategies which we elaborate on in~\secref{subsec:selectivePreciseIntervenstion}.

\section{Field Monitoring} 
\label{subsec:fieldMonitoring}
One of the key benefits of \bbot\ as a platform is that it can monitor the state of the field while traversing it, this supplements its key function as a weeding platform.
We demonstrate the potential of \bbot\ for field monitoring by illustrating how the extra localization sensors can enhance the existing tracking algorithms in our prior work~\cite{halstead2021crop}.
This approach used Mask-RCNN to provide instance-based segmentation, species-level information (e.g. crop and weed species) and the viewable surface area. 
This approach included a tracking-via-segmentation technique that outlined the benefit of of spatial matching operator, coined dynamic radius (DR), over a pixel-wise version, referred to as intersection over union (IoU).
This technique also exploited re-projection between frames using wheel odometry and camera parameters.
This enabled more accurate tracking of objects in the scene, however, wheel odometry is often prone to errors.
Using the extra GPS sensors available on \bbot\ has the potential to increase performance by re-projecting more accurately between frames. 

In this paper we demonstrate that the extra localization sensors on \bbot\ can be used to enhance the performance of field monitoring.
In particular, we fuse the available odometery and GPS information with an EKF~\cite{wei2011intelligent}.
This algorithm recursively estimates the state of a non-linear system in an optimal way.
Furthermore, adding local source of motion estimation can considerably reduce the risk of outage due to lack of proper satellite observations~\cite{ahmadi2021towards}.

In our prior work~\cite{halstead2021crop} we concentrated on sweet pepper in a horticultural setting and sugar beet in arable farmland.
In both cases the objects we aim to detect are somewhat robust to external influences such as weather conditions.
However, some of the weed species witnessed in the arable farmland were grasses and their accurate localization proved difficult.
In this paper we further outline the ability for our approach to be crop agnostic by performing monitoring on a novel grass crop dataset consisting of corn.
Corn has a long leaf structure which makes it susceptible to weather conditions resulting in a difficult crop to localize, due to this, in~\secref{subsec:cropWeedMonitoring} we outline our performance using both the pixel-wise and spatial matching criteria from~\cite{halstead2021crop}.

\begin{figure}[!t]
    \vspace{2mm}
	\begin{tabular}{cc}
	    \hspace{-3mm}
        \includegraphics[width=0.49\linewidth]{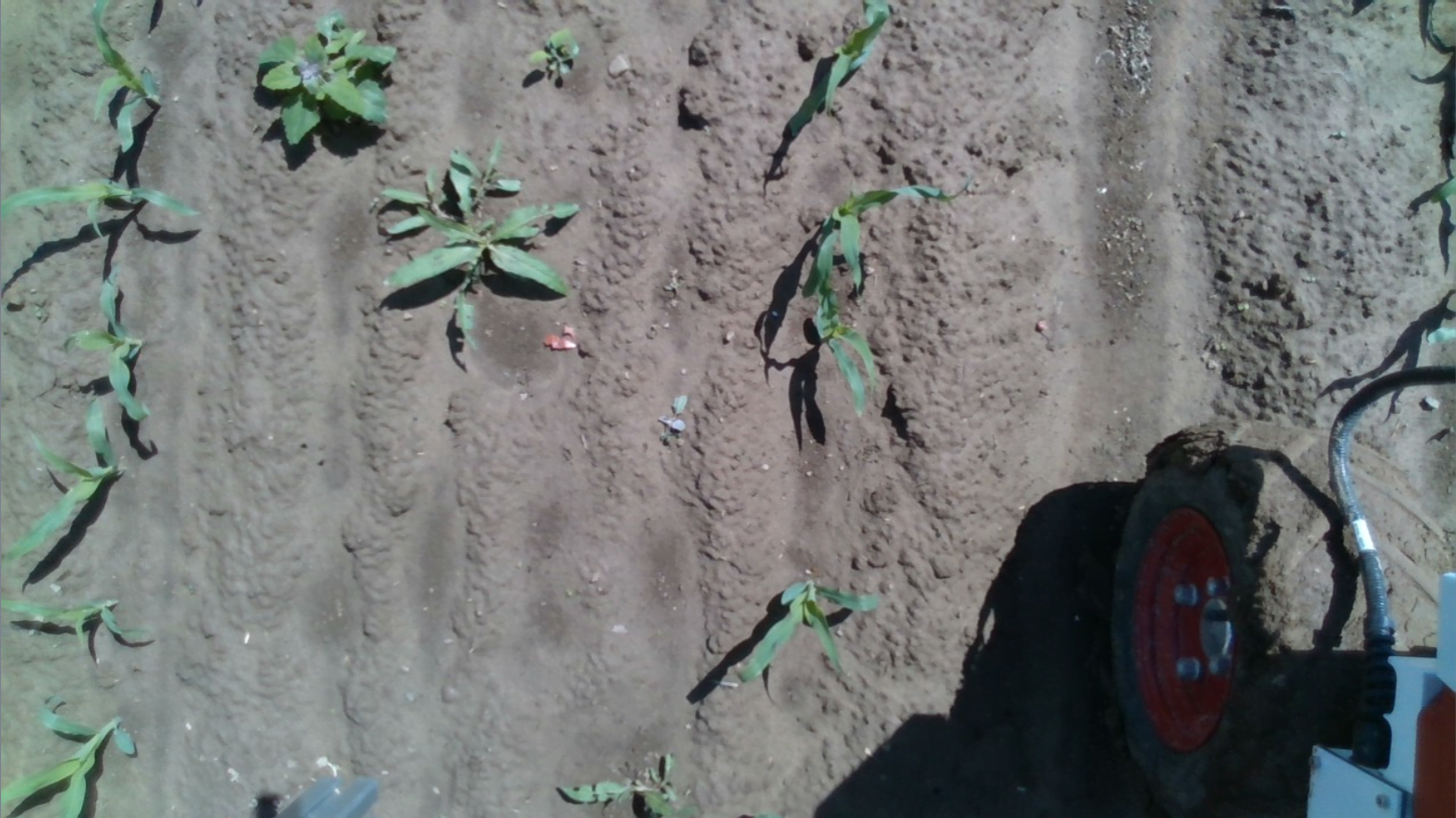}
        \includegraphics[width=0.49\linewidth]{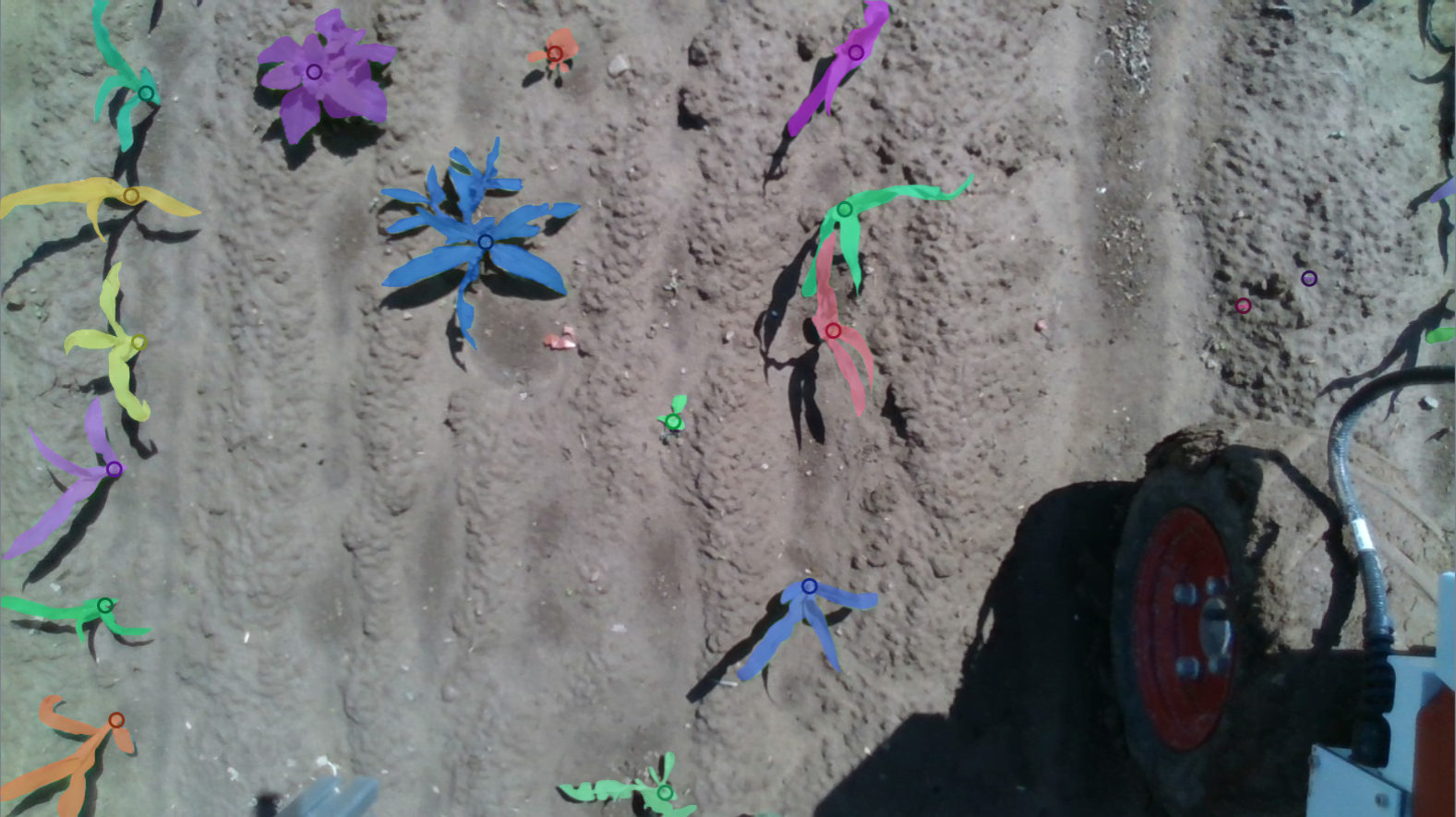}
    \end{tabular}
    \caption{Example image of the \textit{CN20} dataset. 
    Left is the original images and right is the same scene with instance-based annotations with unique color for each plant.}
	\label{fig:datasetSample}
	\vspace{-7mm}
\end{figure}

This corn data set (CN20) was acquired using \bbot\ from a phenotyping field at campus Klein-Altendorf (CKA) of the University of Bonn.
The data was captured using an Intel RealSense D435i sensor with a nadir view of the ground in front of the robot and resolution of $1280\times720$ with a frame-rate of $15Hz$. 
The non-overlapping training, validation, and evaluation data includes RGB-D frames for $170$, $43$ and $70$ images respectively.
This data comes from six different rows providing unique crop and weed distributions due to the non-homogeneous growth stage of the weeds.
In total there are nine different categories of weeds containing a total of $2566$ and $1261$ instances of crop and weeds respectively. 
The data is annotated to include instance based pixel-wise segmentation, bounding boxes and stem locations of each instance in Coco format~\cite{lin2014microsoft}.
\figref{fig:datasetSample} shows an example annotated image of CN20 dataset.

\section{Selective Precise In-Field Intervention} 
\label{subsec:selectivePreciseIntervenstion}

\bbot\ is equipped with a novel weeding tool design enabling high precision plant-level field interventions. 
It consists of a set of replicated linear actuators and is controlled via the intervention controller unit consisting of several components which are elaborated in the following sections.
We briefly explain the conceptual design of the weeding tool, its requirements and operation assumptions.
Then, we introduce our method for managing targets in the work-space of the weeding tools.
Finally, we elaborate on path planning strategies used for controlling intervention heads in action.



\subsection{Plant-level Treatment In Field}
\label{subsec:plantLevelTreatment}

\begin{figure}[!t]
	\centering
    \vspace{2mm}
	\includegraphics[width=1.0\linewidth]{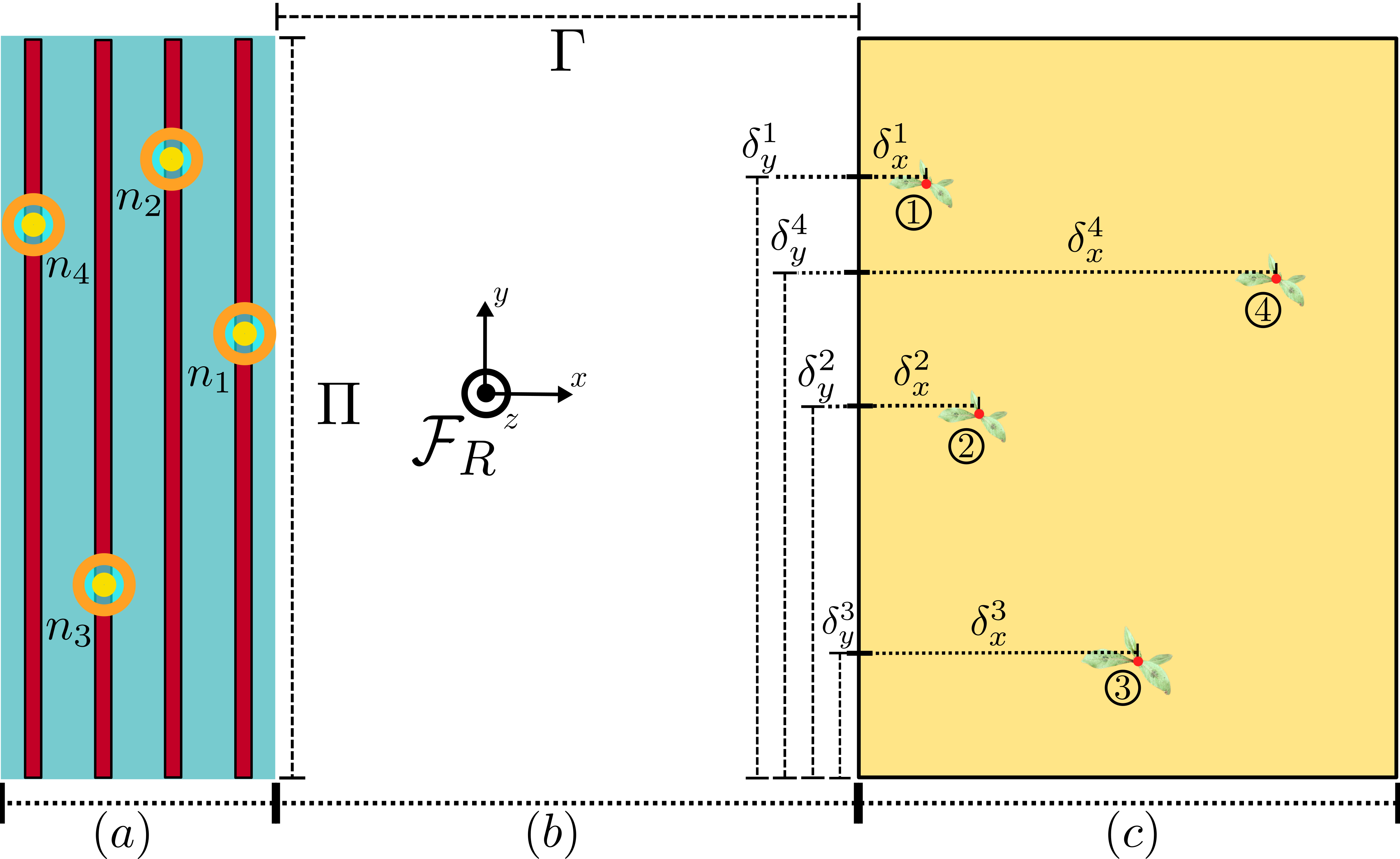}
	\vspace{-2mm}
	\caption{Visualization of Kinematic model of weeding tool work-space (a) the weeds detected in the viewable area of the camera $C_{detect}$ (c); and (b) the gap between two regions.}
	\label{fig:kinematicsWeeding}
	\vspace{-6mm}
\end{figure}

We assume the robot moves along a crop-row with constant speed $\gamma$.
Consequently, intervention is time-critical and must respect the spatial ordering of the weeds.
There is a constant gap ($\Gamma$) between the tools and the area sensed by the camera ($C_{detect}$).
Similar to~\cite{bawden2017robot} we assume weeds are uniformly distributed in the field with density $\lambda$ weeds/$m^2$. 
Hence, using a Poisson process we can explain the distance between the weeding implement and individual weeds by accounting for the arrival rate of $\eta =\lambda \times \Pi$.
We use the motion along the $x$-axis of the robot frame $\mathcal{F}_R$ to explain the weeds interval distance ($\delta_x$), visualized in ~\figref{fig:kinematicsWeeding}.
This can be shown using the following probability density function,
\begin{equation}
    \label{eq:weedsIntervaldistance}
    f(\delta_x) = \lambda \Pi e^{-\lambda \Pi \delta_x},
\end{equation}
also the location of weeds on the $y$ axis can be represented via a uniformly distributed random variable $y$ as,
\begin{equation}
    \label{eq:yPDF}
	f(y) =\left\{\begin{array}{cc}
	    \frac{1}{\Pi} & for  \ \ 0 \leq y \leq \Pi \\
	     0 & otherwise
	\end{array}\right..
\end{equation}
To engage the $i$-th intervention head with the $j$-th weed it has to traverse,
\begin{equation}
    \label{eq:yDistance}
    \delta^{ij}_{y} = | \textit{h}_{i} - \textit{n}_{j} |, 
    \text{ where }  0 \leq \delta^{ij}_{y} \leq \Pi ,
\end{equation}
where $\textit{h}_{i}$ is the current position of $i$-th intervention head and the $\textit{n}_{j}$ denotes to the position of the $j$-th weed.
Therefore, the probability of visiting the $j$-th weed with $i$-th intervention head can be calculated with,
\begin{equation}
    \label{eq:weedingProb}
    \textit{P}_{ij} = \textit{P}\left( \dfrac{\gamma}{\vartheta}  <   \dfrac{\delta_{x}}{\delta_{y}} \right),
\end{equation}
where $\vartheta$ denotes to the maximum velocity of linear axes.
We assume all targets are detected by the time they reach the bottom edge of the camera's viewable area.
%

\subsection{Target-Space Management}
\label{subsec:targetSpaceManagement}

Planning the motion of each intervention head must be done prior to targets entering the weeding tool work-space. 
In the proposed workflow, the intervention controller node receives the detected targets at time $t + \tau_d$ where $\tau_d$ is the time required for detection in the monitoring node.
The monitoring node provides plant specific information like: plant category, pixel-wise segmentation, estimated area, and the bounding box.
Furthermore, we estimate plant centers based on provided bounding box in the scene.  
This information is then used in the target-space management step to assign targets to the intervention heads.
Based on this the next step finds the best motion plan for each intervention head by maximizing the number of targets that are visited (sprayed).

Let $\mathcal{H}$ denote the number of independent controllable intervention heads and $\mathcal{N}$ be the number of targets that appear under the robot. 
We use a uni-directional constrained node-graph to model the targets-space.

\begin{figure}[!t]
	\centering
	\vspace{2mm}
	\includegraphics[width=\linewidth]{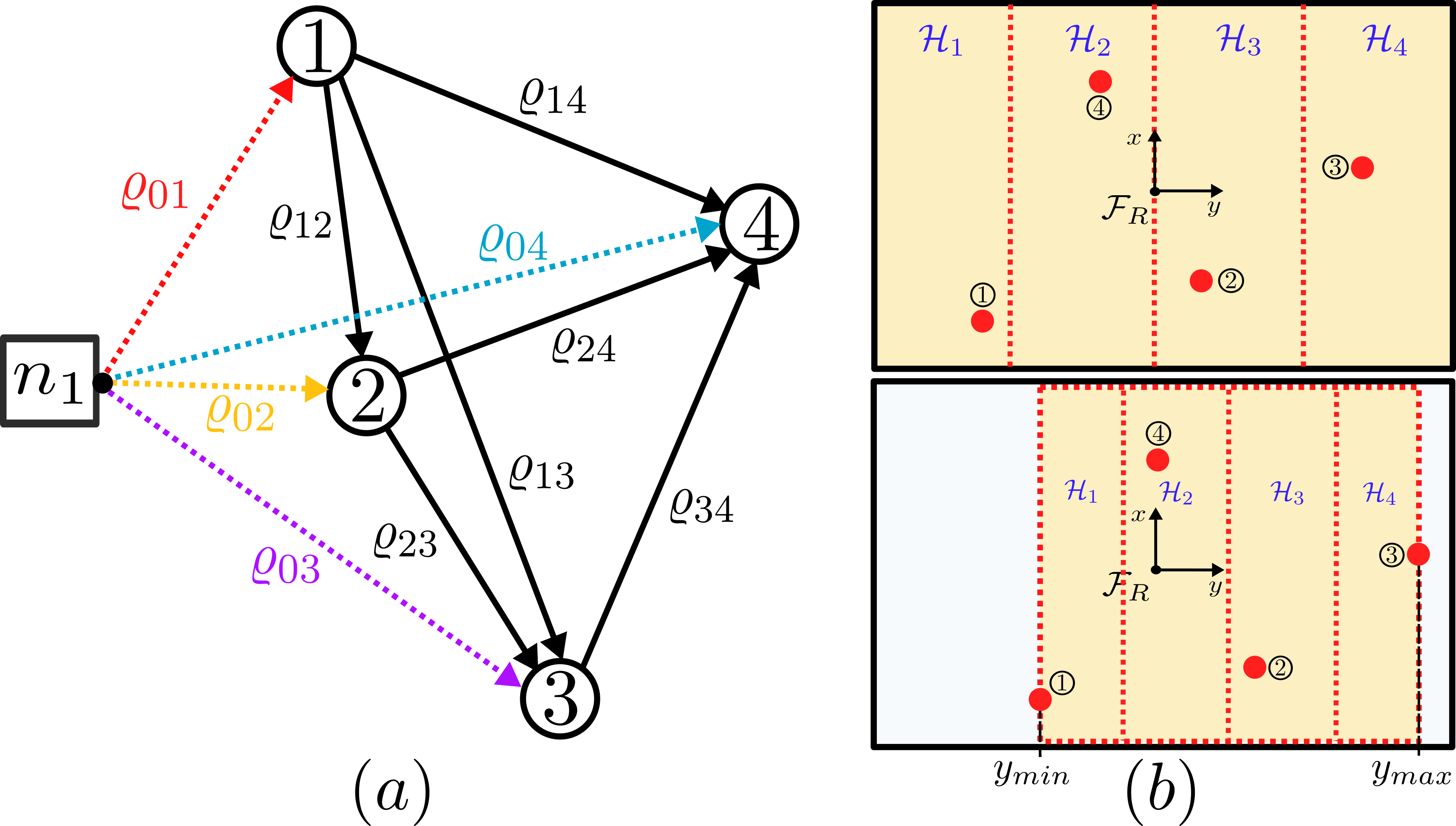}
	\caption{In (a) an example unidirectional constrained node-graph is presented where the numbered circles represent weeds (targets) for one intervention tool $n_{1}$ with different possible start paths (in different colors).
	In (b) we present a visualization of static work-space sub-division (top) and dynamic work-space sub-divisions (bottom).}
	\vspace{-6mm}
	\label{fig:nodeGraph}
\end{figure}

To obtain the global spatial order of targets in a segment we use the $\delta_x$ of each weed (see \figref{fig:kinematicsWeeding}).
In~\figref{fig:nodeGraph}(a), each node (circle) shows a weed along with the connecting path between nodes $j$ to $k$ represented with a uni-directional link (arrow) $l_{jk}$.
The link $l_{jk}$ exists if, node $j$ geometrically is located after node $k$ in the 3D world frame  $\mathcal{F}_w$ in the direction of motion.
Furthermore, the link $l_{jk}$ is associated with an inter-weed cost $\varrho_{jk}$ based on the distance of nodes $j$ and $k$ and a property denoting motion probability of $\textit{P}_{jk}$ based on~\eqref{eq:weedingProb}.
We calculate inter-weed costs using the top-right of the cost-matrix~$\mathcal{G}_{\mathcal{N} \times \mathcal{N}}$ (to respect the weeds spatial order).

%
There are $\mathcal{H}$ independent interventions heads and so multiple plans which can lead to the same number of targets being visited (sprayed).
To solve this problem, we consider the weeds as a sets of targets detected in one location, this motivates us to assign intervention targets to the $\mathcal{H}$ heads as either distance-based or work-space division-based assignments.

\begin{enumerate}
    \item \textbf{\textit{Distance-based Target Assignment (D)}:}
        In this approach, target $j$ gets assigned to the laterally closest intervention head along the sliding direction ($y$-axis). 
        This means, selected intervention head $i$ has the least motion required to reach the weed $j$.
        The lateral distance between heads and weeds are defined based on 2D euclidean distance between projection of intervention head's position on ground plane and weed positions on same plane w.r.t the $\mathcal{F}_w$ frame.
        
    \item \textbf{\textit{Static Work-space Division-based Target Assignment (SD)}:}
        In this method, we divide the work-space of weeding tool to $\mathcal{H}$ sub-sections of width $\Pi/\mathcal{H}$ meters.
        Hence, each intervention head is only responsible for engaging with weeds laying within it's sub-work-space as shown in~\figref{fig:nodeGraph}(b)-top.
        
    \item \textbf{\textit{Dynamic Work-space Division-based Target Assignment (DD)}:}
        In this model, for each new set of detected weeds we first determine the minimum region of intervention defined by $y_{min}$ and $y_{max}$ (see Fig.~\ref{fig:nodeGraph}~(b)). 
        The minimum region of intervention is then divided into $\mathcal{H}$ equal sub-regions.
        This process assists in optimizing the planning for weed engagement by potentially reducing the area any one tool has to cover.
\end{enumerate}
%

\subsection{Intervention Heads Route Planning}
\label{subsec:routePlanning}

Here we address how to plan $\mathcal{H}$ independent efficient routes.
The planned routes must guide intervention heads through all their assigned targets while minimizing the chance of missing any target.
This has to take into account the the prior knowledge of an intervention head's position, robot linear speed as well as the limits of speed and acceleration of linear axes.

The planning approach generates~$m$ potential trajectories~$\Vec{\mathbf{T}}=[\Vec{T}_0,\dots,\Vec{T}_m]$ for each intervention head.
Each trajectory $\Vec{T}_i$ is an ordered list of length $q$ consisting of weed positions which can be visited.
To obtain $\Vec{\mathbf{T}}$ we use the following approaches:
\begin{enumerate}  
    \item \textbf{\textit{Brute-Force}}:
    In this case we compute all possible routes by finding the permutation of all nodes in the graph (without considering the direction of links).
    Then the routes with the lowest cost and maximum success rate will be selected from all predicted routes. 
    \item \textbf{\textit{Open Loop Traveling Salesman Planning}}:
    This approach, termed OTSP, is a variant of the classic travelling salesman problem where the agent must visit all nodes of a graph once without making a loop back (Hamilton loop) to the start node~\cite{chieng2014performance}.
    To solve this we use an approach similar to $n$OTSP where the agent only needs to visit $n$ nodes in the graph, however, in our problem setting we aim to maximize the number of visited nodes while considering other important criteria like cost and success rate.
    We use our constrained uni-directional node-graph representation as a base for solving $n$OTSP using dynamic programming.
\end{enumerate}

The optimal trajectory for each intervention head is obtained by considering two criteria: number of nodes successfully visited and the total movement of the predicted trajectory.
In every trajectory, we calculate the number of nodes that satisfy~\eqref{eq:weedingProb} to determine if a node can be successfully visited.
This gives us an updated set $\Vec{\mathbf{T}'}$ which only consists of nodes in the trajectories which are feasible.
From this updated set $\Vec{\mathbf{T}'}$ we then calculate the movement cost-matrix~$\mathcal{G}$,
\begin{equation}
    \label{eq:trajectoryCost} 
    \mathcal{G}(\Vec{T}'_i) = \sum_{j=0}^{q-1} (n_j - n_{j+1})^2.
\end{equation}
After this process, the trajectory from $\Vec{\mathbf{T}'}$ with the maximum number of successfully vised nodes is passed to the intervention controller.
In the case multiple trajectories successfully visit the same number of nodes, the trajectory which also minimizes the movement cost will be passed to the intervention controller.

\section{Experimental Evaluations}
\label{sec:exp}



\subsection{Field Monitoring} 
\label{subsec:cropWeedMonitoring}

We evaluate the performance improvements to field monitoring techniques when using a range of localization sensors.
Our previous work~\cite{halstead2021crop} is enhanced by fusing the wheel odometry and GNSS, which are available on the platform, using EKF. 
With this information, we are able to improve the average normalized absolute error (NAE) across the three CN20 evaluation rows using our dynamic radius (DR) tracker from $8.3\%$ to $3.5\%$, more than halving the NAE.
This is a considerable improvement as the absolute error of the field tracking system is less than $5\%$ NAE.
We attribute this performance improvement to better localization information supplied through multiple localization sensors and sensor fusion.
To fully appreciate the impact of sensor fusion using EKF we also explore the improvements for a pixel-wise matching tracker (IoU).
This is much simpler tracking approach, although susceptible to leaf movement due to wind, is greatly improvement with the NAE improving from $71.7\%$ to $25.7\%$ with the incorporation of multiple localization sensors and sensor fusion.
Finally, in our experiments we were consistently able to achieve $R^2$ scores greater than $0.95$, showing that our tracker was linearly correlated with the ground truth.
These results demonstrate the importance of multiple localization sensors and the considerable positive impact they can have on existing field monitoring techniques.

\subsection{Weeding Planning Performance} 

To evaluate the performance of the weeding tool planning explained in~\secref{subsec:selectivePreciseIntervenstion}, we use our native python simulator (\secref{subsec:simulation}) specifically designed for this purpose. 
We investigate the performance of our approach using two different types of crop-row models: simulated rows of crops/weeds and real plant distributions predicted from our monitoring systems for both SB20 and CN20 datasets evaluation rows.
We initially evaluated the performance of the Brute-Force planner against the $n$OTSP planner and found the results were similar in trivial cases with a limited number of targets ($\mathcal{N}\leq10$).

The computational expense of the Brute-Force method increases almost exponentially when there are more than four targets as it has a run-time complexity of $O(n!)$ in comparison with $O(n^22^n)$ of graph-based $n$OTSP.
This is a significant problem for weeding applications where real-time performance is important.
In the case where we see only ten weeds in the planning region, the Brute-Force approach requires $3.7$s compared to $n$OTSP which requires only $266\mu$s.
This is a prohibitive quality of the Brute-Force approach and for our two weeding experiments we employ the $n$OTSP technique.


\subsubsection{Planning on Simulated Crop-Rows}

In our first experiment, we evaluate the performance of different weed densities and a different number of linear axes.
This evaluation uses the simulation environment introduced in Sec.~\ref{subsec:simulation}, allowing us to control the experimental parameters.
For our platform hyper-parameters, we keep a constant robot speed of $\gamma=0.5$ and set the velocity of the linear actuators to $\vartheta=5$.
The field parameters are set to $3$ crops-rows in a single lane with a length of $20$m.
To fully analyze the performance of our approach we vary the weed density such that $\lambda\,=\left[3, 5, 10, 20, 40\right]$ represents the weeds per $m^2$.
Finally, to outline the benefit of having multiple linear axes we show the performance for $\mathcal{H}\,=\left[1,2,4,8\right]$.

\begin{figure}[!t]
    \centering
    \vspace{2mm}
    \includegraphics[width=1.0\linewidth]{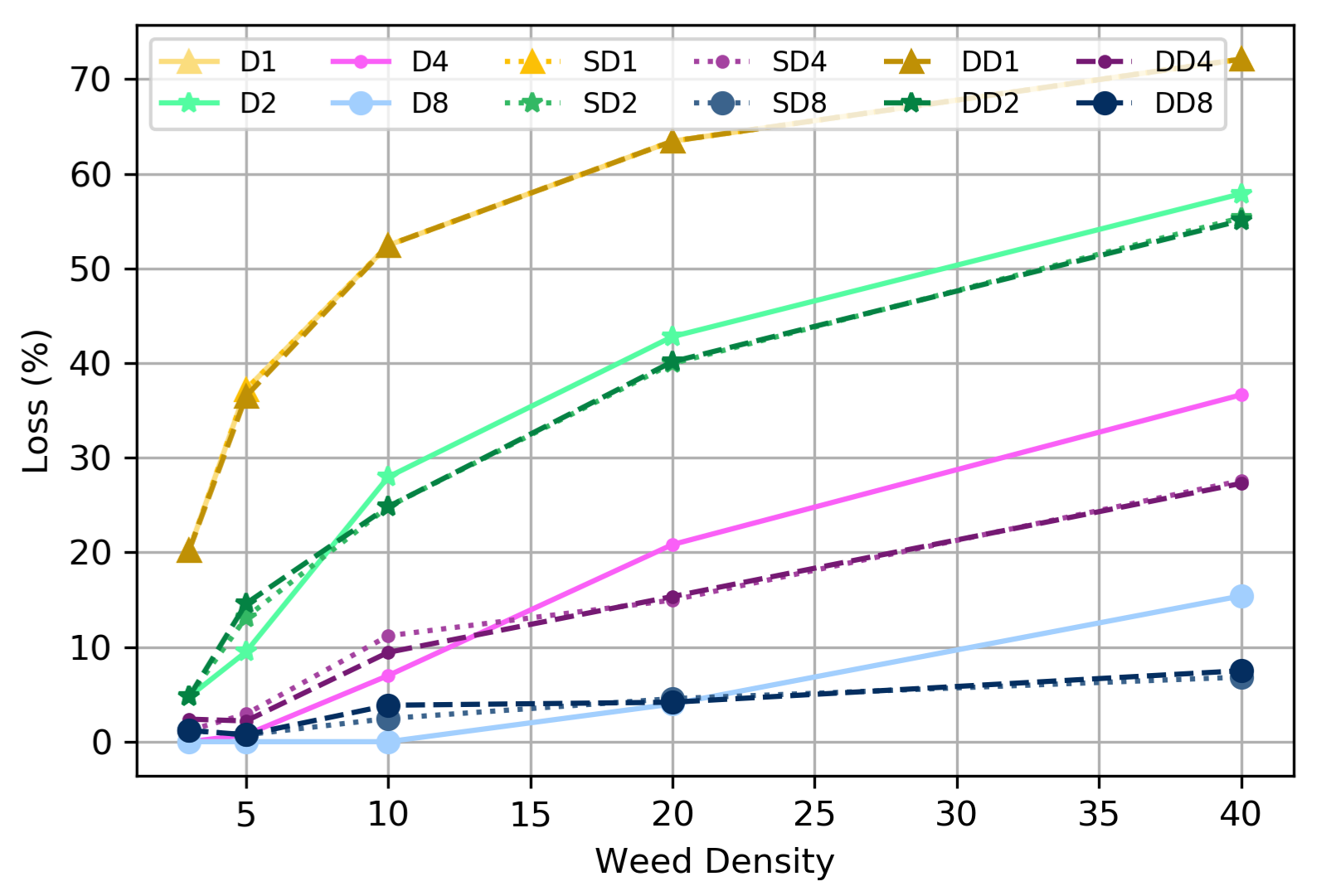}
    \vspace{-6mm}
    \caption{
    The results of the $n$OTSP weeding planner. The loss represents the number of missed weeds.
    We evaluate 5 different weed density models (x-axis); and three different planning types: Dist.-Based (D), Dyn.-Div. (DD), and Sub-Div. (SD); and intervention heads $\mathcal{H}=1, 2, 4, 8$, yellow, green, purple and blue, respectively.}
    \label{fig:simulationPerformace}
    \vspace{-6mm}
\end{figure}

The results for this simulated experiment are summarized in Fig.~\ref{fig:simulationPerformace}, where we provide the comparison of weed density to the percentage of missed targets.
From this figure it is evident that increasing the number of linear axes has an obvious impact on results, with $8$ axes performing better than all others.
We see from this that even with a distribution of $40$ weeds (the hardest case) the worst performing $8$ axes system, Distance-Based, achieves a loss of $\sim{15\%}$.


Overall, the Distance-Based approach routinely performs worse as the weed density increases, this is particularly evident as the number of heads is increased.
This will be further evaluated in the next evaluation, but as the distribution grows the intervention heads need to travel further to meet the demands of the planner, this movement can negatively impact the capacity for the intervention head to reach the next weed.
Finally, the two division-based methods appear to perform at a commensurate level across all weed densities in our simulations.
We attribute this to the wide distribution of weeds negating the impact of the dynamic approach, meaning, weeds generally appear across the entire lane rather than concentrated in a specific region.
Overall, this shows the validity of our planning methods for weed intervention in a uniformly distributed pattern.

\subsubsection{Planning on Real Crop-Rows}
To further evaluate the performance of our three planning approaches, we use real weed distributions captured in the fields within the weeding simulator.
This information is obtained from the crop monitoring approach outlined in~\secref{subsec:fieldMonitoring} and aggregated into a simulated row.
We perform this on the evaluation rows for both the SB20 and CN20 datasets where we have three different weed distributions: low (CN20), medium (SB20-S1), and high (SB20-S2).

The results of our three planning approaches are displayed in~\tabref{tab:realPlanning}, in this experiment we only use four intervention heads as this accurately evaluates \bbot\ performance.
\tabref{tab:realPlanning} displays two metrics, first, the percentage of missed weeds, and second, the mean and standard deviation of the distance moved by the axes.

\begin{table}[!b]
    \vspace{-6mm}
	\centering
	\caption{The rate of Loss ($\%$) and average traveled distance ($m$) of interventions heads in real-world weeding scenarios.}
	\resizebox{\columnwidth}{!}{%
	\begin{tabular}{l | cc|cc|cc}
	\toprule
	 & \multicolumn{2}{c|}{\textbf{Sub-Div. (SD)}}  & \multicolumn{2}{c|}{\textbf{
	 Dist.-Based (D)}} & \multicolumn{2}{c}{\textbf{Dyn.-Div. (DD)}} \\
	 & (\%) & (m) & (\%) & (m) & (\%) & (m) \\\hline	 
	 \midrule
	CN20        & \textbf{0.0}  & 2.7$\pm$0.2    & 4.3    & 2.7$\pm$2.8   & 3.4          & 4.0$\pm$3.9  \\
    SB20-S1     & \textbf{0.0}  & 1.4$\pm${0.2}  & 2.3    & 1.5$\pm${0.6} & \textbf{0.0} & 1.0$\pm${0.8}\\
    SB20-S2     & \textbf{11.9} & 10.1$\pm${0.9} & 19.8   & 4.7$\pm$3.5   & 13.5         & 5.0$\pm$1.8  \\
    \bottomrule
    \end{tabular}}
	\label{tab:realPlanning}
\end{table}

For the percentage of missed weeds, we see that the SB20-S2 with a high density of weeds is the most difficult to intervene on, this is somewhat expected due to the heavy distribution of weeds.
However, the simple Sub-division approach outperforms the other two approaches for this distribution which is unlike the shown results in~\figref{fig:simulationPerformace} for a density of $10$ per $m^2$.
The poorer performance of the two other approaches can be attributed to the distribution of the weeds in a real crop row.
We performed the Chi-squared test~\cite{turhan2020karl} on the sections of the data from the rows and found that the distribution of fields was not uniform, hence increasing the complexity.
This performance is mirrored through the other rows, where the static Sub-division approach achieves higher scores.
However, once we reduce the weed density (CN20, SB20-S1) we are able to achieve a percentage of missed weeds close to zero for all approaches.

Our final evaluation is based on the movement requirements of the different planning approaches, where a value closer to zero is desired.
While static Sub-division achieved the best performance for the percentage of missed weeds we see the negative aspect of this approach here.
Both the dynamic division and distance-based approaches can achieve considerably better results in the traveled distance, in the case of SB20-S2 both other approaches move half as much as the static Sub-division approach.
We believe that in the future this work can be used to provide baseline information about robotic path planning for weeding applications.
This includes potential improvements by jointly minimizing the total distance traveled (of the intervention heads) and the loss.

\section{Conclusion}
\label{sec:conc}

In this paper, we introduced \bbot\ a crop monitoring and weeding platform specifically designed for European phenotyping fields.
We showed how \bbot\ can greatly improving crop monitoring, reducing the NAE from $8.3\%$ to $3.5\%$, by combining the extra localization sensors it has available.
Furthermore, we present a novel arrangement of weeding tools mounted on linear actuators along with three associated target planning approaches.
These target planning approaches were evaluated in simulated environments which are able to mirror real-world weed distributions by using the results from our field monitoring approach.
Experiments on these simulated real-world fields demonstrated the validity of our proposed work-space division techniques and showed that they led to significantly less movement ($10m$ compared to $5m$) when compared to distance-based target assignment.
Finally, we found that while the real-world data was generally uniform in future any planning systems should be evaluated on real-world weed distributions.
\section*{Acknowledgements}

This work was funded by the Deutsche Forschungsgemeinschaft (DFG, German Research Foundation) under Germany’s Excellence Strategy - EXC 2070 – 390732324.

\bibliography{references}
\bibliographystyle{IEEEtran}

\end{document}